\documentclass{llncs}

\usepackage[pdftex]{graphicx}
\usepackage{amsmath,amsfonts,amssymb}
\usepackage{multirow}
\usepackage{rotating}
\usepackage{floatrow}

\usepackage{subfigure}
\usepackage{color}
\usepackage{longtable}
\usepackage{tikz}

\definecolor{myred}{rgb}{.8,.0,.0}

\newcommand{\auc}[2]{\mathbf{AUC}^{(#1)}_{#2}}
\newcommand{\roc}[2]{\mathbf{ROC}^{(#1)}_{#2}}

\newcommand{\clasfnum}[0]{22}

\begin{document}

\title{Characterizing Multiple Instance Datasets}


\author{Veronika Cheplygina$^{1,2}$, David M. J. Tax$^2$}
\institute{$^1$ Biomedical Imaging Group Rotterdam, Erasmus Medical Center, The Netherlands\\
$^2$ Pattern Recognition Laboratory, Delft University of Technology, The Netherlands
}

\maketitle
\thispagestyle{empty}

\begin{abstract}
In many pattern recognition problems, a single feature vector is not sufficient to describe an object. In multiple instance learning (MIL), objects are represented by sets (\emph{bags}) of feature vectors (\emph{instances}). This requires an adaptation of standard supervised classifiers in order to train and evaluate on these bags of instances. Like for supervised classification, several benchmark datasets and numerous classifiers are available for MIL. When performing a comparison of different MIL classifiers, it is important to understand the differences of the datasets, used in the comparison. Seemingly different (based on factors such as dimensionality) datasets may elicit very similar behaviour in classifiers, and vice versa. This has implications for what kind of conclusions may be drawn from the comparison results. We aim to give an overview of the variability of available benchmark datasets and some popular MIL classifiers. We use a dataset dissimilarity measure, based on the differences between the ROC-curves obtained by different classifiers, and embed this dataset dissimilarity matrix into a low-dimensional space. Our results show that conceptually similar datasets can behave very differently. We therefore recommend examining such dataset characteristics when making comparisons between existing and new MIL classifiers.

The datasets are available via Figshare at \url{https://bit.ly/2K9iTja}.    
\end{abstract}

\section{Introduction}\label{sec:intro}

Images portraying several objects, text documents covering a range of topics or molecules with conformations with different chemical properties are all examples of data, where a single example (image, document, molecule) cannot always be faithfully represented by a single feature vector. Representing each part (object in an image, paragraph in a text document, molecule conformation) of an example by a single feature vector preserves more information about the example, but requires a finer level of annotation, which is not always available. To deal with such problems, supervised learning has been extended to multiple instance learning (MIL): a learning scenario where examples are sets (\emph{bags}) of feature vectors (\emph{instances}), but where labels are available only for bags. Originally, the goal in MIL was to classify previously unseen bags, however MIL classifiers which are able to classify instances have also received a lot of attention because of their ability to be trained with only coarse annotations.

Since the introduction~\cite{dietterich1997solving} of MIL in 1997, many classifiers have been proposed in the literature. A typical strategy in comparisons is to evaluate on the early benchmark problems (Musk~\cite{dietterich1997solving}, Fox, Tiger and Elephant~\cite{andrews2002multiple}) as well as a number of larger sources, such as MIL adaptations of Corel~\cite{chen2006miles} image datasets, or Newsgroups~\cite{zhou2009multi} text classification problems, which consist of 20 datasets each. Usually one of the following strategies is used when choosing datasets for a comparison: (i) targeting a particular application, such as image classification, and choosing few sources with many datasets per source (ii) choosing diverse datasets, for example by choosing many sources, with a few datasets per source, and/or choosing datasets with different characteristics, such as dimensionality.

A potential pitfall in choosing datasets this way is that, while they may seem diverse to a human observer, this may not be the case for a classifier, and vice versa. For example, in a related study on characterizing standard datasets~\cite{DuiPekTax2004}, Duin et al. show that when changing dataset size and dimensionality for three different problems, some modified datasets remain similar (in dataset space) to their original versions. This is very important for the types of conclusions that can be drawn from an empirical comparison on a ``observer-diverse'' or ''observer-similar'' set of problems. For example, a classifier which performs well on a ``observer-diverse'' set of problems, may in fact only be suitable for problems in a small area of the dataset space. On the other hand, a classifier that is very good in one area of the dataset space, but not performing well on ``observer-similar'' problems might delay (or even prevent) the paper from being published.

In this paper we review a large number of problems that have been used as benchmarks in the MIL literature. We propose to quantify the dataset similarity based on the behavior of classifiers, namely by comparing the ROC curves, or the area under the ROC curves, that different classifiers obtain on these datasets. Our results show that conceptually similar datasets can behave quite differently. When comparing MIL classifiers, we therefore recommend not choosing datasets based on the application (images, text, and so forth) or on the dataset properties (bag size, dimensionality), but on how differently existing classifiers perform on these datasets.


\section{Multiple Instance Learning}

In multiple instance learning\cite{dietterich1997solving}, a sample is a set or \emph{bag} $B_i$ of feature vectors $\{\mathbf{x}_i^1, \ldots, \mathbf{x}_i^{n_i}\}$. Each bag is associated with a label $y_i \in \{0, 1\}$, while the instances are unlabeled. Often assumptions are made about the existence of instance labels $\{z_i^k\}$, and their relationships to $y_i$. The standard assumption is ``a bag is positive if and only if it has a positive instance'', but over the years, more relaxed assumptions have been explored~\cite{foulds2010review}. The positive instances are often called concept instances, and an area in the feature space with positive instances is often referred as ``the concept''.

\begin{figure}
\centering
\includegraphics[width=0.8\textwidth]{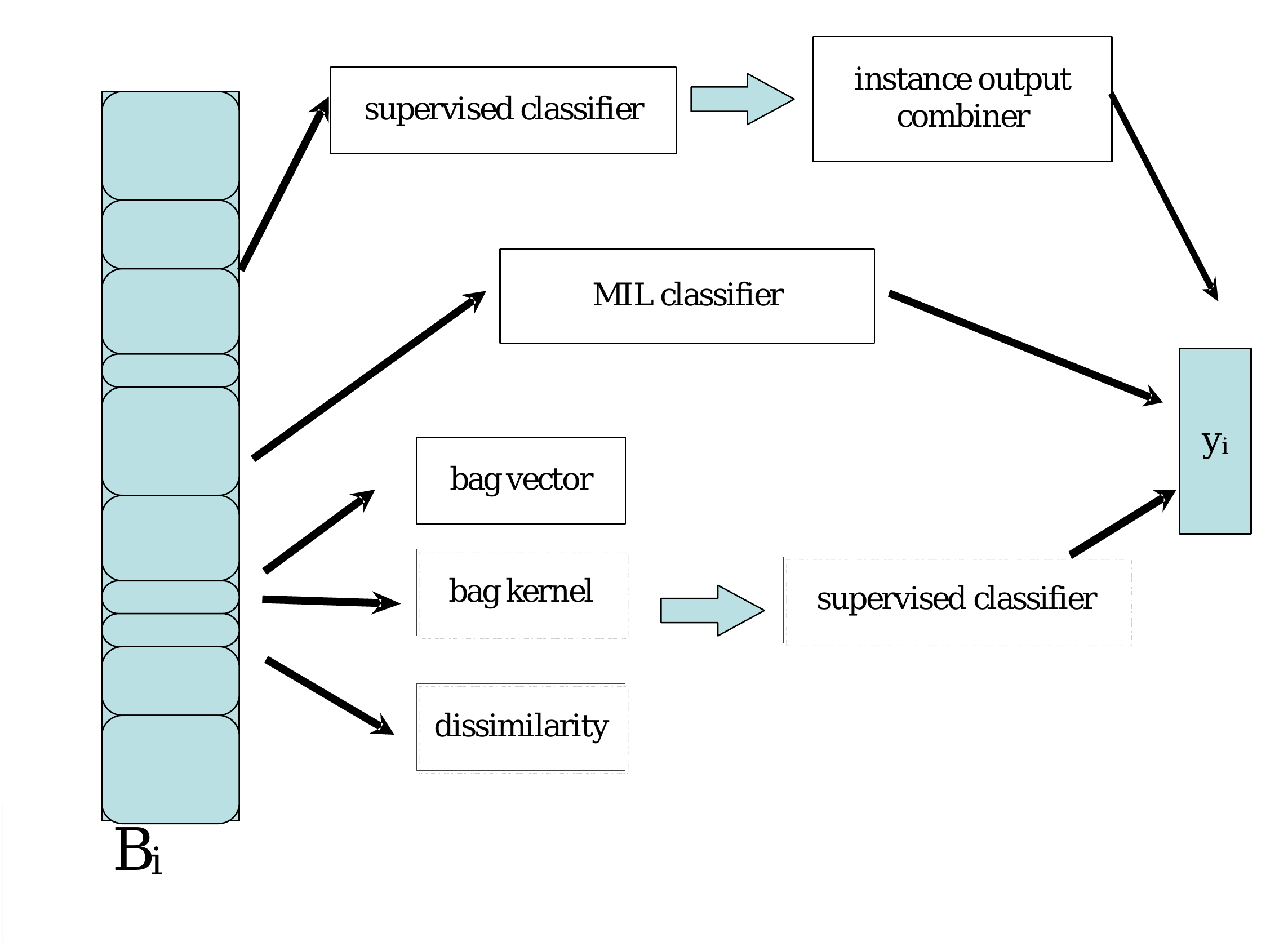}
\caption{A dataset with bags $B_i$ of varying number of instances, and three general approaches how to arrive at bag labels $y_i$}
\label{fig:classifiers}
\end{figure}

Originally, the goal in MIL is to train a classifier $f_B$, which can label previously unseen bags. Globally, this can be achieved either by (1) training an instance classifier $f_I$, which relies on the assumptions about the instance and bag labels, and defining $f_B$ by combining outputs of $f_I$, or (2) training $f_B$ directly, by defining a supervised representation of the bags, or by distance- and kernel-based methods. We call these approaches instance-level and bag-level approaches, respectively. These approaches, which are also summarized in Fig.~\ref{fig:classifiers}, are as follows:

\begin{description}

\item[Supervised classifier] By assuming that all the instances in a bag share the bag's label, a supervised classifier can be trained. A test bag is classified by combining the outputs of its instances. We call this approach simpleMIL.

\item[MIL classifier] By using the standard MIL assumption of a concept (or a generalization thereof), an instance classifier can be trained, which is consistent with the training bag labels. Examples used in this paper are Diverse Density~\cite{maron1998framework}, EM-DD~\cite{zhang2001dd}, MILBoost~\cite{zhang2005multiple} and miSVM~\cite{andrews2002support}. In Diverse Density the concept is explicitly modeled as an ellipsoidal region around one location. This location, and the dimensions of the ellipsoid, are optimized by maximizing the data likelihood. The concept should have high ``diverse density'': high density of positive instances but low density of instances from negative bags. EM-DD is an expectation-maximization algorithm which searches for the concept. The expectation step selects the most positive instance from each bag according to the current estimate for the concept, and the maximization step updates the concept by maximizing the diverse density. The miSVM classifier extends the regular SVM by searching not only for the optimal decision boundary, but also for the instance labels, which, given the decision boundary, are consistent with training bag labels.

\item[Bag vector, kernel or dissimilarity] This approach converts the bag into an alternative representation before training a supervised, bag-level classifier. Examples used in this paper are Citation-kNN~\cite{wang2000solving}, bag statistics~\cite{gartner2002multi}, bag-of-words, MILES~\cite{chen2006miles} and MInD~\cite{cheplygina2015multiple,tax2011bag}. Citation-kNN defines a bag distance based on the number of ``referencing'' nearest neighbors, and the number of ``citing'' neighbors, and applies a nearest neighbor classifier. The other approaches represent each bag by a single feature vector, and apply a supervised classifier. The representation is absolute (instance statistics per bag) or relative, in terms of similarities to instance clusters (bag of words), instances in the training set (MILES), and bags in the training set (MInD).
\end{description}

A complete overview of MIL classifiers can be found in~\cite{amores2013multiple}.

\section{Datasets}\label{sec:datasets}

In this section we describe the datasets we use in the experiments. These include 6 artificial datasets and 34 real-life datasets from 13 groups. For the artificial datasets, we use
three datasets where only a number of concept instances are informative, and three datasets where all instances are informative. For the real-life datasets, different groups represent different sources of data. In some cases, different datasets from the same source are obtained by splitting up a multi-class problem into different one-against-all problems. For such groups, we use a small number of datasets per group to make sure that the influence of each group is not too large. The complete list of datasets is shown in Table~\ref{tab:mildata}.

\subsection{Artificial datasets}

\begin{description}

\item[Gaussian] For the positive bags, instances are drawn from the positive concept Gaussian centered around (7,1), and a random set of instances is drawn from a background Gaussian distribution around (0,0). For the negative bags, the instances are drawn from the background distribution.

\item[Maron] Instances are randomly drawn from a uniform distribution in a unit square. For positive bags, one instance is also drawn from the 5 interval in the center of the square. This dataset is originally defined in \cite{maron1998framework}.

\item[Concept] Instances are randomly drawn from 4 Gaussian distributions with centers [+2,-2], [-2,+2] or [-2,-2].  For positive bags, at least one instance is also drawn from [+2,+2].

\item[Difficult] Both positive and negative instances are drawn from elongated Gaussian distributions, that differ in mean in only the first feature.

\item[Rotated] The instances are drawn from an elongated Gaussian distributions. Instances from positive bags are drawn from a slightly rotated version of the negative instance distribution.

\item[Widened] The instances are drawn from two Gaussian distributions. For positive bags, this Gaussian is slightly wider than for negative bags.
\end{description}

\subsection{Real-life datasets}

\subsubsection{Biology}

\begin{description}

\item[Musk] are molecule activity prediction problems. Each bag is a molecule, each instance is one of that molecule's conformations. The molecule is active if at least one of its conformations is active.

\item[Mutagenesis] is a molecule activity prediction problem~\cite{srinivasan1995comparing}. Each bag is a molecule and each instance is a pair of atoms in that molecule, described by their chemical properties.

\item[Protein] is a problem of predicting whether a protein belongs to a family of TrX proteins~\cite{tao2004svm}. A bag is protein, and an instance is part of that protein's sequence, represented by its molecular and chemical properties.
\end{description}

\subsubsection{Images}

\begin{description}
\item[Corel] are scene classification problems~\cite{chen2006miles}. Each bag is an image, each instance is a patch of that image. The images depict scenes of a beach, historical buildings, and so forth. Using the original 20 classes, 20 datasets are generated using the one-against-all approach.

\item[SIVAL] are image classification problems~\cite{rahmani2005localized}. The images show a particular object (such as an apple) from different perspectives and in front of different backgrounds. Datasets are generated by the one-against-all approach.

\item[Fox, Tiger, Elephant] are image classification problems~\cite{andrews2002multiple}. The positive images show the respective animal, the negative images are selected randomly from other (more than just these three) classes.

\item[Breast] is an image classification problem~\cite{kandemir2014empowering}. A bag is a tissue microarray image and an instance is a patch. The task is to predict whether the image is malignant (positive) or benign (negative).

\item[Messidor] is an image classification problem~\cite{kandemir2015computer}. A bag is an eye fundus image and an instance is a patch. The task is to predict whether the image is of a subject with diabetes (positive) or a healthy subject (negative).
\end{description}

\subsubsection{Text}

\begin{description}
\item[Web] are text classification problems~\cite{zhou2005multi}. A bag is a webpage, and an instance is a webpage that the original page links to. The goal is to predict whether to recommend a particular webpage to a user based on the content of the linked pages. The data in each of the datasets are the same, but the labels are different for each user.

\item[Newsgroups] are text classification problems~\cite{zhou2009multi}. A bag is a collection of newsgroup posts, each described by frequencies of different words. A positive bag for a category contains 3\% of posts about that category, whereas negative bags contain only posts about other topics.

\item[Biocreative] is a text classification problem~\cite{ray2005learning}. A bag is a biomedical text and an instance is paragraph in the document. The task is to predict whether the text should be annotated as relevant for a particular protein.

\end{description}

\subsubsection{Other}

\begin{description}
\item[Harddrive] is a problem of predicting harddrive failures~\cite{murray2006machine}. Each bag are time series (instance = time point) of different measurements of hard drives, and each bag is labeled with whether a failure has occured or not.

\item[Birds] are concerned with classifying whether a particular bird is present in a sound recording~\cite{briggs2012acoustic}. A bag is a recording's spectrogram, an instance is a segment of that spectrogram. Datasets are generated by the one-against-all approach.

\end{description}


\begin{table}[ht]
\begin{center}
\begin{tabular}{l*{6}{c}}
& \multicolumn{6}{c}{2} \\
                  & + bags & -- bags & Features & Total inst & Min & Max \\ 
 \hline 
Musk 1                                       & 47 & 45 & 166 & 476 &  2 & 40 \\
Musk 2                                       & 39 & 63 & 166 & 6598 &  1 & 1044 \\
Gaussian-MI                       & 50 & 50 &  2 & 692 &  5 &  9 \\
Maron-MI                                     & 50 & 50 &  2 & 1000 & 10 & 10 \\
MI-concept                                   & 10 & 10 &  2 & 126 &  5 &  8 \\
Difficult-MI                                & 10 & 40 &  2 & 352 &  5 &  9 \\
Rotated-MI                                   & 30 & 30 &  2 & 1359 & 15 & 29 \\
Widened-MI                                   & 30 & 30 &  2 & 1259 & 15 & 29 \\
Corel African                                & 100 & 1900 &  9 & 7947 &  2 & 13 \\
Corel Beach                                  & 100 & 1900 &  9 & 7947 &  2 & 13 \\
Corel Historical                             & 100 & 1900 &  9 & 7947 &  2 & 13 \\
Corel Buses                                  & 100 & 1900 &  9 & 7947 &  2 & 13 \\
Corel Dinosaurs                              & 100 & 1900 &  9 & 7947 &  2 & 13 \\
Corel Elephants                              & 100 & 1900 &  9 & 7947 &  2 & 13 \\
Corel Food                                   & 100 & 1900 &  9 & 7947 &  2 & 13 \\
Sival AjaxOrange                             & 60 & 1440 & 30 & 47414 & 31 & 32 \\
Sival Apple                                  & 60 & 1440 & 30 & 47414 & 31 & 32 \\
Sival Banana                                 & 60 & 1440 & 30 & 47414 & 31 & 32 \\
Sival BlueScrunge                            & 60 & 1440 & 30 & 47414 & 31 & 32 \\
Web recomm. 1                                & 17 & 58 & 5863 & 2212 &  4 & 131 \\
Web recomm. 2                                & 18 & 57 & 6519 & 2219 &  5 & 200 \\
Web recomm. 3                                & 14 & 61 & 6306 & 2514 &  5 & 200 \\
Web recomm. 4                                & 55 & 20 & 6059 & 2291 &  4 & 200 \\
Text(Zhou) alt.atheism                       & 50 & 50 & 200 & 5443 & 22 & 76 \\
Text(Zhou) comp.graphics                     & 49 & 51 & 200 & 3094 & 12 & 58 \\
Text(Zhou) comp.os.ms-windows.misc           & 50 & 50 & 200 & 5175 & 25 & 82 \\
Fox (Andrews)                                & 100 & 100 & 230 & 1320 &  2 & 13 \\
Tiger (Andrews)                              & 100 & 100 & 230 & 1220 &  1 & 13 \\
Elephant (Andrews)                           & 100 & 100 & 230 & 1391 &  2 & 13 \\
Harddrive (positive=non-failed)              & 178 & 191 & 61 & 68411 &  2 & 299 \\
Protein                                      & 25 & 168 &  8 & 26611 & 35 & 189 \\
Mutagenesis easy                             & 125 & 63 &  7 & 10486 & 28 & 88 \\
Mutagenesis hard                             & 13 & 29 &  7 & 2132 & 26 & 86 \\
Birds, target class Brown Creeper            & 197 & 351 & 38 & 10232 &  2 & 43 \\
Birds, target class Winter Wren              & 109 & 439 & 38 & 10232 &  2 & 43 \\
Birds, target class Pacific-slope Flycatcher & 165 & 383 & 38 & 10232 &  2 & 43 \\
Birds, target class Red-breasted Nuthatch    & 82 & 466 & 38 & 10232 &  2 & 43 \\
Biocreative component                        & 359 & 359 & 200 & 13129 &  1 & 53 \\
UCSB Breast cancer                           & 26 & 32 & 708 & 2002 & 21 & 40 \\
Messidor retinopathy                         & 654 & 546 & 687 & 12352 &  8 & 12 \\
\end{tabular}

\end{center}

\caption{List of MIL datasets and their properties: number of positive and negative bags, number of features, number of instances and minimum/maximum number of instances per bag.}
\label{tab:mildata}

\end{table}

\section{Proposed Approach}

To summarize and embed the results of all classifiers on all datasets, we define a distance or similarity between datasets and results.
The most simple representation uses basic metadata about a dataset. These features can be, for instance, the dimensionality, the number of bags, the number of instances, and so forth. When this metadata representation of a dataset $i$ is $M^{(i)}$, the distance between two datasets is easily defined as:

\begin{equation}
D_{meta}(X_i, X_j) = \|M^{(i)} - M^{(j)}  \|.
\label{eq:Dmeta}
\end{equation}

These metadata features are typically not very informative for how classifiers perform on these datasets. For this, the outputs of the classifiers are needed. A standard approach is to compare the predicted labels and count how often two classifiers disagree in their prediction \cite{DuiPekTax2004}. Unfortunately, for MIL problems this approach is not very suitable, because MIL classification problems can have a very large class imbalance (as is visible in the Corel and SIVAL datasets). The alternative is to use the receiver-operating characteristic (ROC) curves instead. An ROC curve shows the true positive rate as a function of the false positive rate. Because the performances on the positive and negative class is decoupled onto two independent axes, class imbalance does not influence the curve.

\begin{figure}[ht]
\begin{tikzpicture}
\centering
\node [anchor=south west] at (0,0) {\includegraphics[width=6cm]{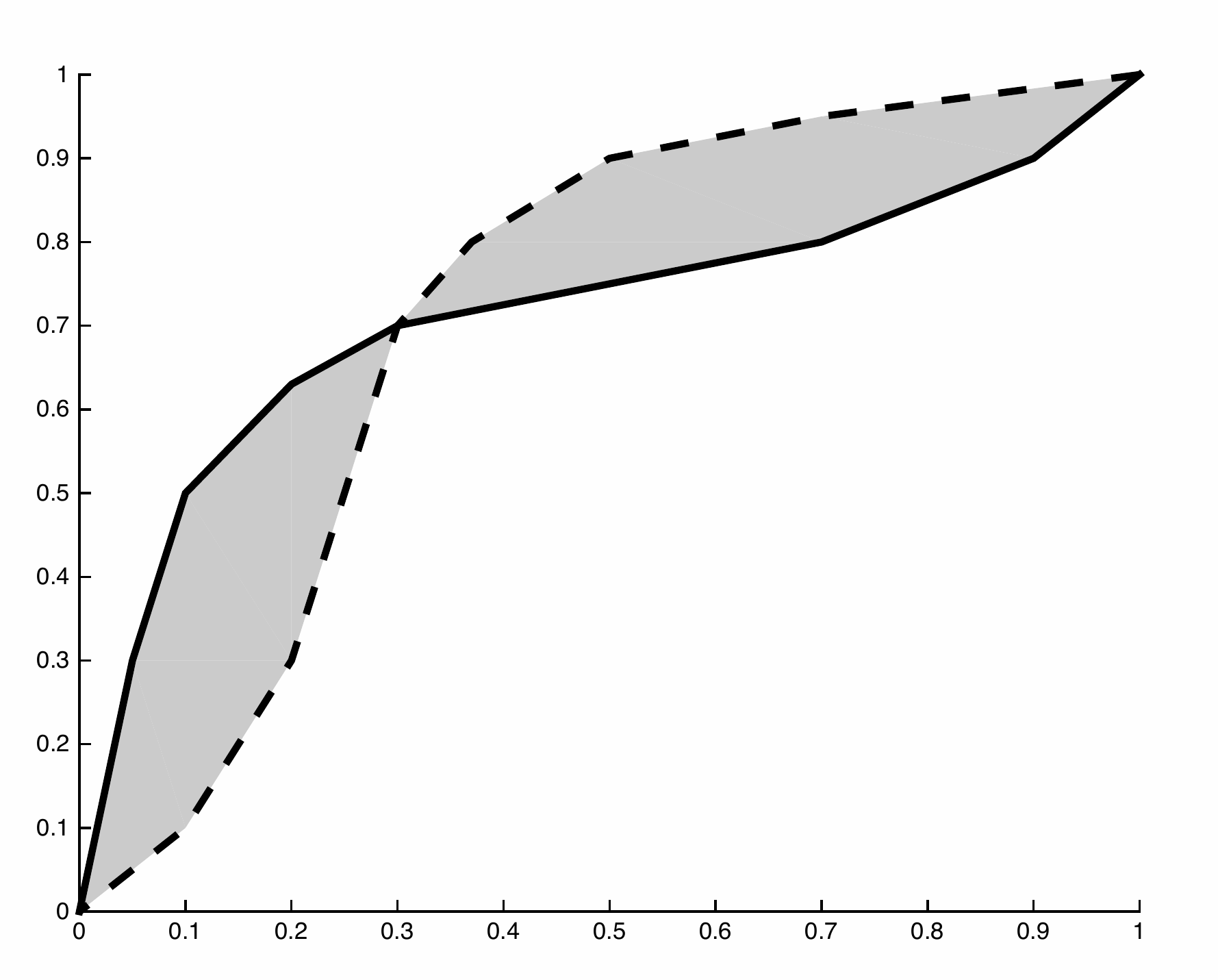}};
\node at (4,3.4) {$\mathbf{ROC}_1$};
\node at (2.3,4.4) {$\mathbf{ROC}_2$};
\end{tikzpicture}
\caption{Two ROC curves $\mathbf{ROC}_1$ and $\mathbf{ROC}_2$ with an equal area under the ROC curve ($\mathcal{A}(\mathbf{ROC})$), but where the two curves differ. The area of the gray region is the area $\mathcal{A}(\mathbf{ROC}-\mathbf{ROC})$. }
\label{fig:diffROC}
\end{figure}

A drawback of the ROC curve is that it is not straightforward to compare two different curves. We choose two different approaches to do this. The first approach is to summarize each ROC curve by its area under the curve (AUC), and compare the different AUCs. This may be suboptimal, because two ROC curves can have an identical AUCs, while their shapes may still be very different. This is illustrated in Fig. \ref{fig:diffROC}, where two curves with equal AUCs are shown, $\mathbf{ROC}_1$ and $\mathbf{ROC}_2$ (solid line and dashed line, respectively). In order to differentiate between these two curves, a second approach is used. Here the area of the difference between the two ROC curves is used as the distance between the curves. This is indicated by the gray area in Fig. \ref{fig:diffROC}.

Let the ROC curve of classifier $k$ on dataset $i$ be $\roc{i}{k}$, and
the AUC performance of classifier $k$ on dataset $i$ be $\auc{i}{k} = \mathcal{A}(\roc{i}{k})$.
In the first approach, the distance between datasets $X_i$ and $X_j$ is defined as:
\begin{equation}
D_{auc}(X_i,X_j) = \|\auc{i}{} - \auc{j}{}\|
\label{eq:D1}
\end{equation}
where $\auc{i}{}$ is the vector of AUC performances, i.e. all performances of all classifiers on dataset $X_i$:
\begin{equation}
\auc{i}{} = \left[\begin{array}{c}
\mathcal{A}(\roc{i}{1}),\\
\vdots\\
\mathcal{A}(\roc{i}{L})
\end{array}\right]
\end{equation}
In the second approach the area under the difference between two ROC curves is used:
\begin{equation}
D_{roc}(X_i,X_j) = \sqrt{\sum_k \mathcal{A}(\roc{i}{k}-\roc{j}{k})^2 }
\label{eq:D2}
\end{equation}

In the above, we have chosen the Euclidean norm to ensure that differences in the embeddings are caused by the choice of representation of the data, rather than by differences in the (non-)Euclideanness of the distances.

\textbf{Embedding and out-of-sample extension.} Given the distances \eqref{eq:Dmeta}, \eqref{eq:D1} or \eqref{eq:D2}, we embed the datasets using multi-dimensional scaling (MDS) \cite{cox2000multidimensional}. MDS places a 2D vector for each dataset, such that the (Euclidean) distances in the 2D embedding match the given distances as good as possible. To compare a previously unseen dataset $Z$ to the datasets in the embedding, the procedure is as follows. First all classifiers are trained on $Z$ and the resulting ROC curves of the test sets are determined. Then the distances $D_{auc}$ or $D_{roc}$ are computed, and finally the 2D location of the $Z$ is optimized to reproduce the original distances as well as possible.

Other algorithms could be considered for embedding as well. We have briefly experimented with t-SNE~\cite{van2008visualizing}, which had a tendency to position the samples on a uniform grid, failing to reveal structure inside the data. In our experience, this happens when only a few samples need to be embedded. Furthermore, the out-of-sample extension is not as straightforward as for classical scaling approaches~\cite{gisbrecht2012out}.

\section{Experiments}

In the experiments, we aim to demonstrate the embeddings for distances $D_{meta}$, $D_{auc}$ and $D_{roc}$ for the datasets described in Section~\ref{sec:datasets}. For $D_{meta}$, we use 6 features which are displayed in Table~\ref{tab:mildata} and normalize these to zero mean and unit variance. For $D_{auc}$ and $D_{roc}$, we use a set of \clasfnum ~classifiers: simpleMIL, diverse density, EM-DD, MILBoost, Citation k-NN$\times 2$, miSVM$\times 2$, MILES$\times 2$, MIL kernel$\times 3$, bag statistics$\times 3$, bag of words$\times 3$, bag dissimilarity$\times 3$.  The base classifier for simpleMIL, bag statistics, bag of words and bag dissimilarity approaches is the logistic classifier. The different versions per classifier type correspond to different classifier parameters for which we have observed different behaviors in earlier work~\cite{tax2011bag,cheplygina2015multiple}. These performances of these classifiers are available through http://homepage.tudelft.nl/n9d04/milweb/ . 

\begin{figure}[ht]
\centering

\includegraphics[width=0.45\textwidth]{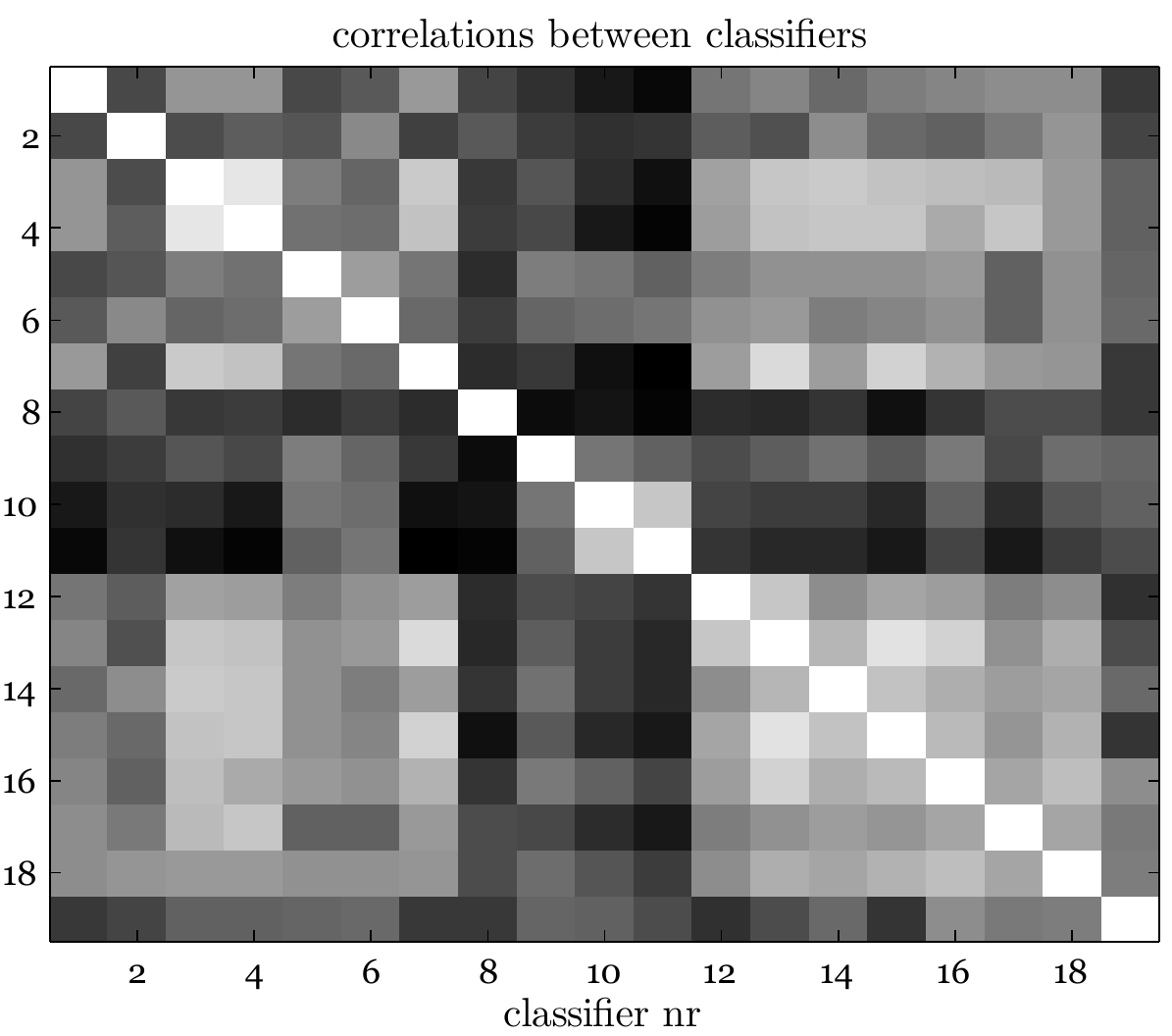}
\includegraphics[width=0.45\textwidth]{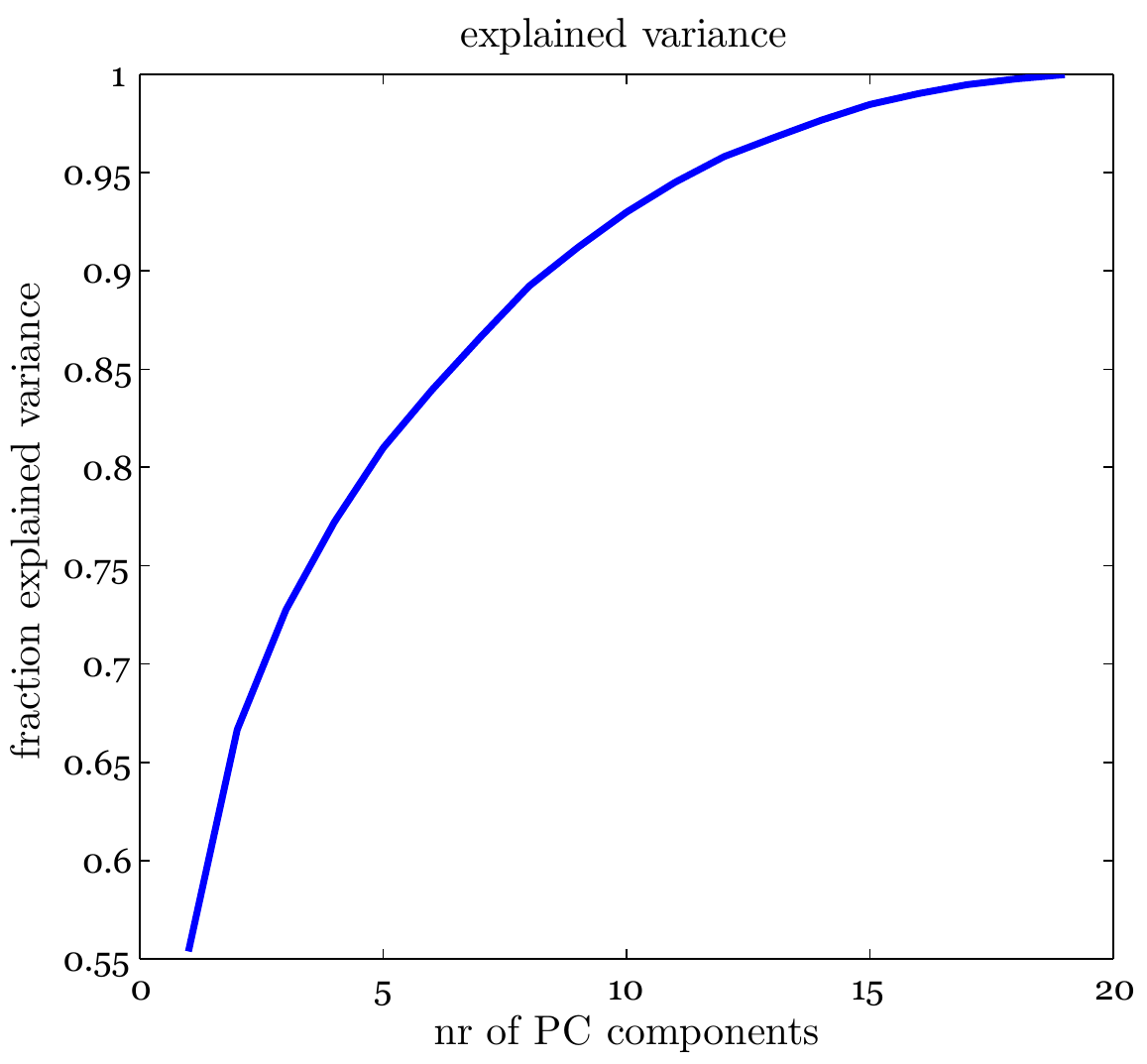}

\caption{Left: Correlations (white = 1, black = 0) between distances ($D_{roc}$) given by each of the \clasfnum ~classifiers. Right: Cumulative fraction of variance explained in the $D_{roc}$ distances between all datasets of \clasfnum ~MIL classifiers.}
\label{fig:var_explained}
\end{figure}

Clearly, the embeddings of $D_{auc}$ and $D_{roc}$ depend on the classifiers which are evaluated. Therefore, we first verify that we are using a diverse set of classifiers. We first create a \clasfnum-dimensional dataset where each feature contains all pairwise distances based a single classifier. We then compute the correlations between the features of this dataset.
We also perform principal component analysis on this data, and compute the cumulative fraction of variance explained by the principal components. The results are shown in Fig.~\ref{fig:var_explained}. The slope of the cumulative fraction of variance suggests that the classifiers are diverse, i.e., if there were two groups of highly correlated classifiers, the slope would be much steeper.

We now compare the embeddings given by the three distances.  $D_{auc}$ and $D_{roc}$ have very similar embeddings, so we show only $D_{roc}$. This means that the situation sketched in Fig.~\ref{fig:diffROC} does not occur very often, i.e. classifiers with similar AUCs also have similar ROC curves.

When comparing $D_{meta}$ and $D_{roc}$ the differences are very large. With $D_{meta}$ some datasets from the same source have exactly the same representation and are on top of each other in the embedding, while the classifiers behave differently on these datasets. Another big difference is in the artificial datasets: these are relatively clustered together in $D_{meta}$, but display drastically different behaviors with $D_{roc}$.

\begin{figure}[ht]
\includegraphics[width=0.4\textwidth]{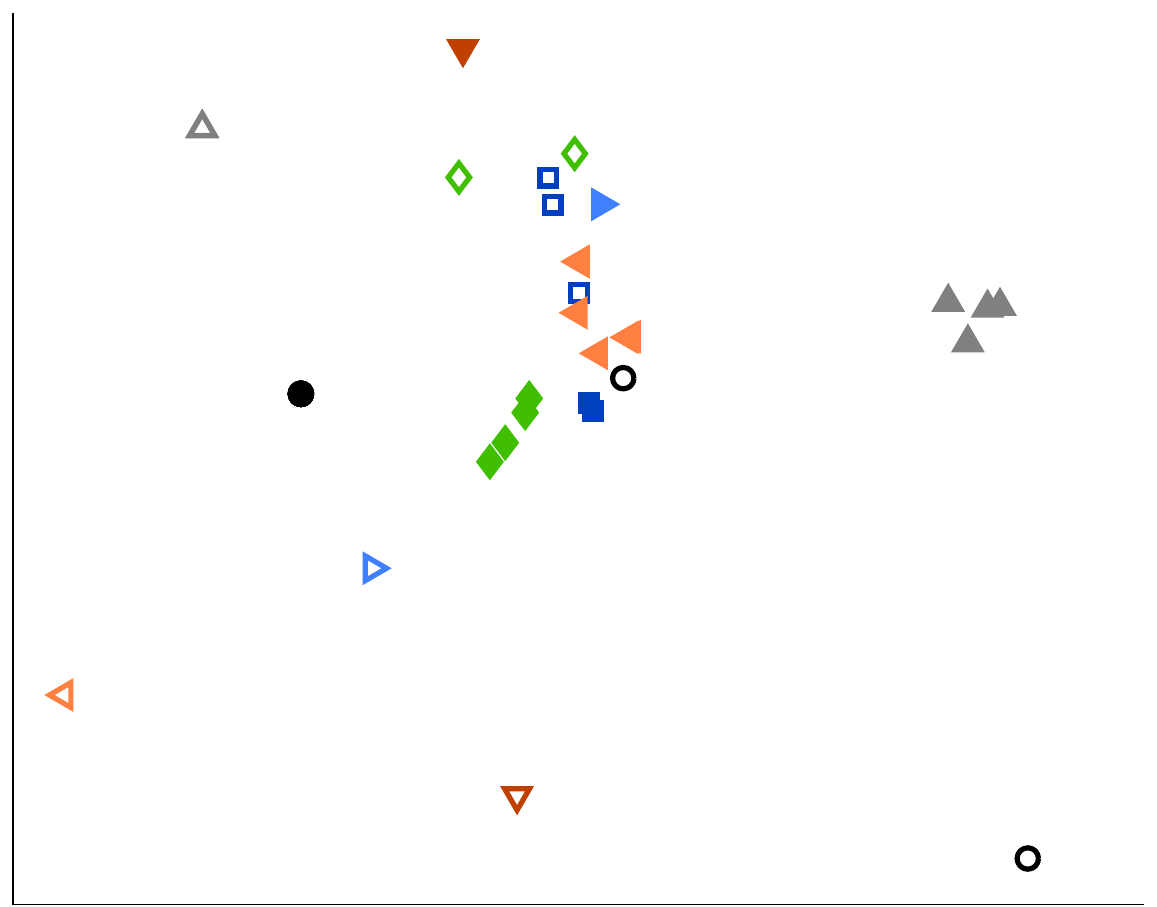}
\includegraphics[width=0.4\textwidth]{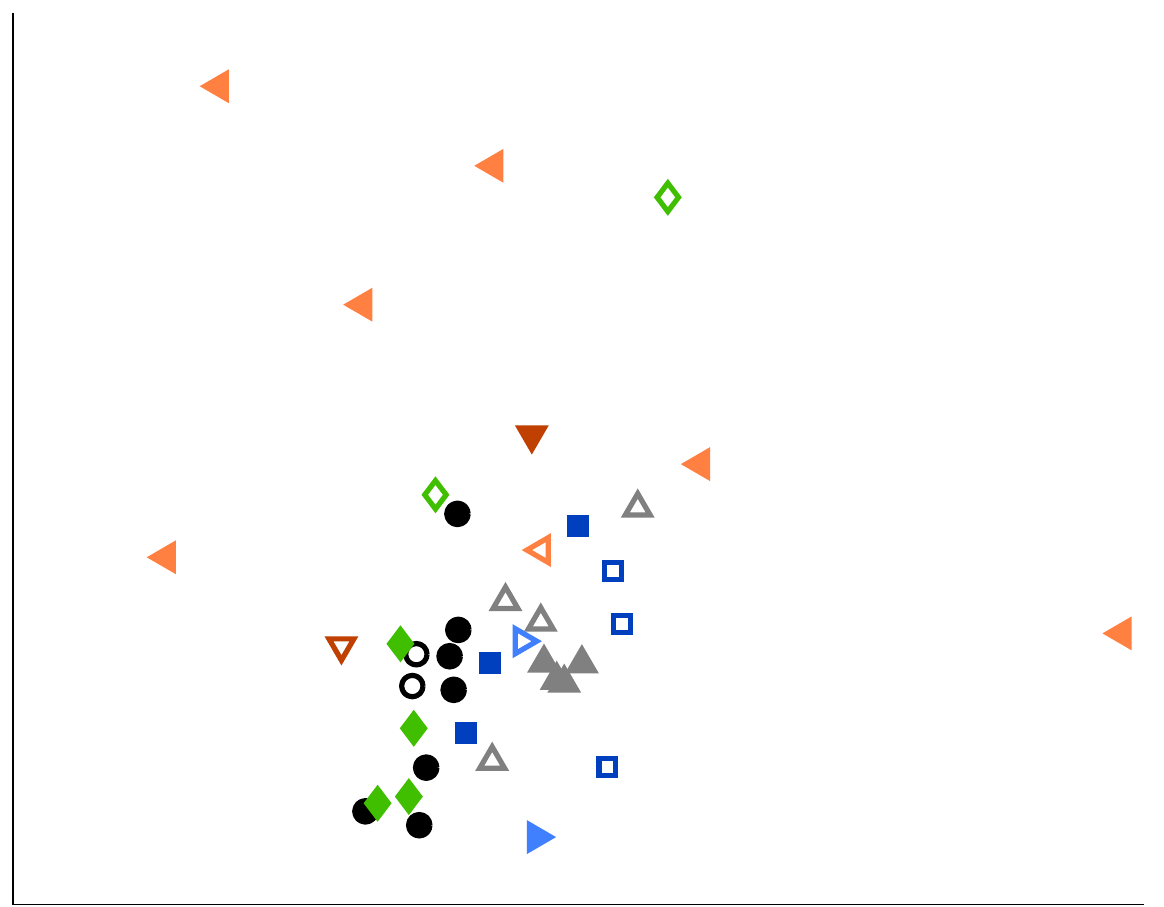}
\includegraphics[width=0.15\textwidth]{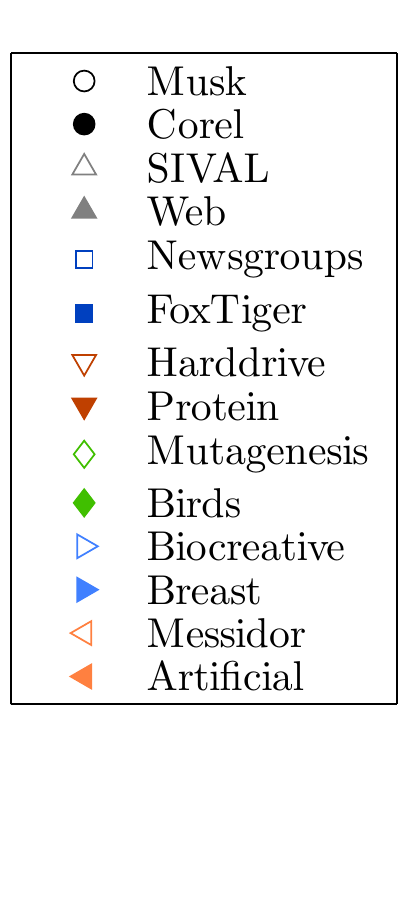}

\caption{Left: MDS embedding of the Euclidean distances between the meta-representations of the datasets. Right: MDS embedding of $D_{roc}$ based on differences of ROC curves.}
\label{fig:alldata}
\end{figure}

We now zoom in more on the $D_{roc}$ embedding. In the Web datasets, the most similar behavior within a dataset group can be observed. For most other dataset groups, we see different behavior of the datasets inside a group. In some situations, such as Birds data, the inside-group variations are smaller than, for example, Corel or SIVAL. This suggests that choosing a different class as the positive class (as is done in the Corel and SIVAL datasets) can change the character of a MIL dataset quite a lot.

A surprising observation is that the artificial datasets are outlier datasets, although they are supposed to be simpler versions of different situations (only one instance is informative, or all instances are informative) encountered in MIL. The differences of the artificial and real data suggest that the real-life datasets may contain a mixture of a concept region (or several concept regions), as well as different background distributions (i.e. negative instances in positive bags are different from negative instances in negative bags). The concept-like artificial datasets are generated such that these background instances are not informative. But in real-life cases, negative instances in positive bags could still be correlated with the bag label. For example, if foxes are photographed in forests more often than other animals, negative instances in a positive bag, i.e. patches of forest in an image of a fox, would still help in classifying the bag as positive.

Another interesting observation is that the datasets which have not been used as benchmarks very often, such as Harddrive, Breast and Biocreative are all quite different from each other. They are also quite different from the more frequently used datasets, such as Musk or Corel. Including these newer datasets in comparisons would therefore be helpful to get a more complete picture of the differences between classifiers.

\begin{figure}[ht]
\includegraphics[width=5cm]{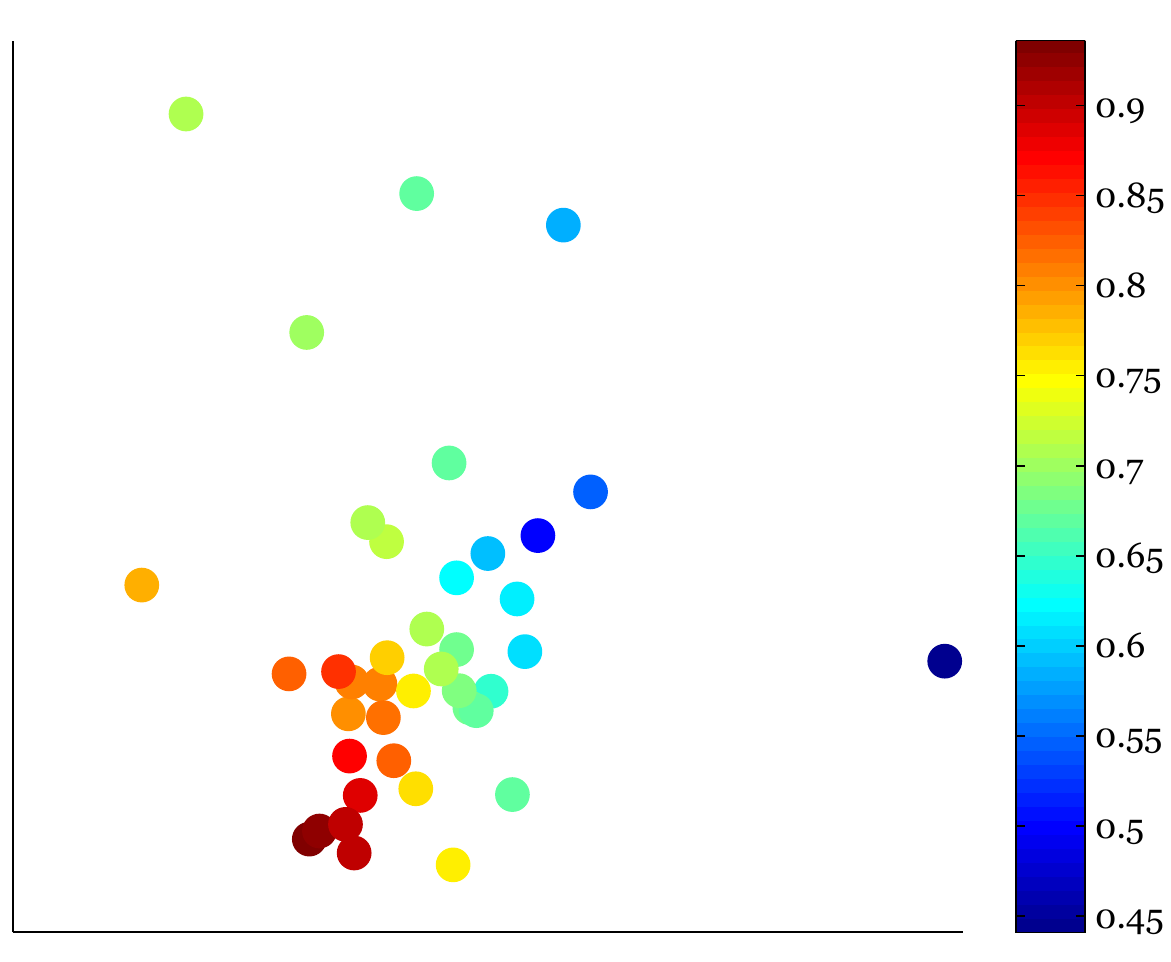}
\includegraphics[width=5cm]{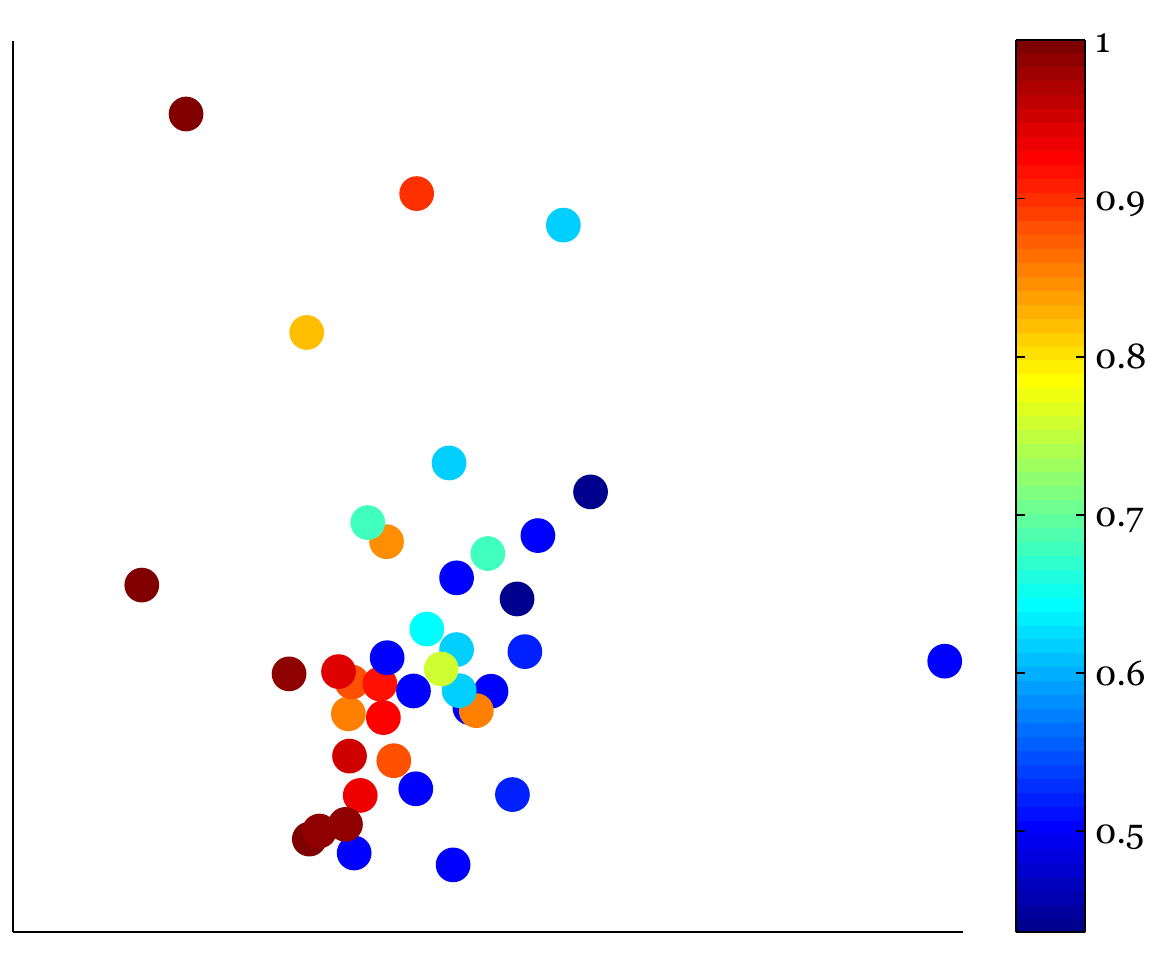}
\caption{Left: the average performance of a dataset, averaged over all MIL classifiers and all crossvalidation folds. Right: the performance of only a concept-based MIL classifier, the EMDD. Low performance is indicated in blue, high performance is indicated in yellow or red.}
\label{fig:heatmaps}
\end{figure}

An attempt can be made to interpret the main variations in the MDS embedding of Fig. \label{fig:alldata}. It appears that the main direction is, not surprisingly, the average performance that the classifiers can achieve on the datasets. In Fig.\ref{fig:heatmaps} again the datasets are shown, embedded by MDS. In the left subplot the datasets are colored by the average performance of all classifiers. The scatter plot suggests that the ``easier'' datasets are in the bottom left. In the right subplot only the performance of the EM-DD classifier is shown. Here it can be observed that datasets on the left tend to have higher performances. It appears that these datasets have a concept present, that is in particular suitable for EM-DD, but also Diverse Density or MILBoost classifiers.

\section{Conclusions}


We proposed to characterize multiple instance learning datasets by quantifying their differences by the differences of ROC curves that different classifiers obtain on these datasets. We have shown that datasets which have similar properties such as the number of bags or instances, can have very different characteristics in terms of classifier behavior. Datasets from the same source, such as datasets derived from a multi-class problem, do not necessarily display similar characteristics. Finally, some datasets which are have not been used in comparisons of MIL classifiers often, behave quite differently from the more frequently used benchmarks. We believe that the proposed approach is useful when deciding which MIL datasets to use in a comparison of classifiers, and in interpreting results obtained by a novel MIL classifier.

A possible extension to the current work is to characterize the datasets by the ranks of the classifiers, rather than the actual performances. Perhaps in such a comparison a more apparent trend between datasets with a concept, multiple concepts, and so forth, would be seen. Another interesting direction of investigation is creating datasets -- artificially, or by subsampling the real-life datasets --  which will fill in the gaps in the dataset space we have investigated.



--------------------------------------------
\bibliographystyle{splncs}
\bibliography{refs}

\begin{thebibliography}{10}
\providecommand{\url}[1]{\texttt{#1}}
\providecommand{\urlprefix}{URL }

\bibitem{amores2013multiple}
Amores, J.: Multiple instance classification: Review, taxonomy and comparative
  study. Artificial Intelligence  201,  81--105 (2013)

\bibitem{andrews2002multiple}
Andrews, S., Hofmann, T., Tsochantaridis, I.: Multiple instance learning with
  generalized support vector machines. In: National Conference on Artificial
  Intelligence. pp. 943--944 (2002)

\bibitem{andrews2002support}
Andrews, S., Tsochantaridis, I., Hofmann, T.: Support vector machines for
  multiple-instance learning. In: Advances in Neural Information Processing
  Systems. pp. 561--568 (2002)

\bibitem{briggs2012acoustic}
Briggs, F., Lakshminarayanan, B., Neal, L., Fern, X.Z., Raich, R., Hadley,
  S.J.K., Hadley, A.S., Betts, M.G.: Acoustic classification of multiple
  simultaneous bird species: A multi-instance multi-label approach. The Journal
  of the Acoustical Society of America  131,  4640 (2012)

\bibitem{chen2006miles}
Chen, Y., Bi, J., Wang, J.: Miles: Multiple-instance learning via embedded
  instance selection. IEEE Transactions on Pattern Analysis and Machine
  Intelligence  28(12),  1931--1947 (2006)

\bibitem{cheplygina2015multiple}
Cheplygina, V., Tax, D.M.J., Loog, M.: Multiple instance learning with bag
  dissimilarities. Pattern Recognition  48(1),  264--275 (2015)

\bibitem{cox2000multidimensional}
Cox, T.F., Cox, M.A.: Multidimensional scaling. CRC Press (2000)

\bibitem{dietterich1997solving}
Dietterich, T.G., Lathrop, R.H., Lozano-P{\'e}rez, T.: Solving the multiple
  instance problem with axis-parallel rectangles. Artificial Intelligence
  89(1-2),  31--71 (1997)

\bibitem{DuiPekTax2004}
Duin, R., Pekalska, E., Tax, D.: The characterization of classification
  problems by classifier disagreements. In: Pattern Recognition, 2004. ICPR
  2004. Proceedings of the 17th International Conference on. vol.~1, pp.
  141--143 Vol.1 (Aug 2004)

\bibitem{foulds2010review}
Foulds, J., Frank, E.: A review of multi-instance learning assumptions.
  Knowledge Engineering Review  25(1), ~1 (2010)

\bibitem{gartner2002multi}
G{\"a}rtner, T., Flach, P.A., Kowalczyk, A., Smola, A.J.: Multi-instance
  kernels. In: International Conference on Machine Learning. pp. 179--186
  (2002)

\bibitem{gisbrecht2012out}
Gisbrecht, A., Lueks, W., Mokbel, B., Hammer, B.: Out-of-sample kernel
  extensions for nonparametric dimensionality reduction. In: Proceedings of
  European Symposium on Artificial Neural Networks (ESANN). pp. 531--536 (2012)

\bibitem{kandemir2015computer}
Kandemir, M., Hamprecht, F.A.: Computer-aided diagnosis from weak supervision:
  A benchmarking study. Computerized Medical Imaging and Graphics, in press
  (2015)

\bibitem{kandemir2014empowering}
Kandemir, M., Zhang, C., Hamprecht, F.A.: Empowering multiple instance
  histopathology cancer diagnosis by cell graphs. In: Medical Image Computing
  and Computer-Assisted Intervention--MICCAI 2014, pp. 228--235. Springer
  (2014)

\bibitem{van2008visualizing}
Van~der Maaten, L., Hinton, G.: Visualizing data using t-sne. Journal of
  Machine Learning Research  9(2579-2605), ~85 (2008)

\bibitem{maron1998framework}
Maron, O., Lozano-P{\'e}rez, T.: A framework for multiple-instance learning.
  Advances in neural information processing systems pp. 570--576 (1998)

\bibitem{murray2006machine}
Murray, J.F., Hughes, G.F., Kreutz-Delgado, K.: Machine learning methods for
  predicting failures in hard drives: A multiple-instance application. Journal
  of Machine Learning Research  6(1),  783 (2006)

\bibitem{rahmani2005localized}
Rahmani, R., Goldman, S.A., Zhang, H., Krettek, J., Fritts, J.E.: Localized
  content based image retrieval. In: International Workshop on Multimedia
  Information Retrieval. pp. 227--236. ACM (2005)

\bibitem{ray2005learning}
Ray, S., Craven, M.: Learning statistical models for annotating proteins with
  function information using biomedical text. BMC bioinformatics  6(Suppl 1),
  S18 (2005)

\bibitem{srinivasan1995comparing}
Srinivasan, A., Muggleton, S., King, R.D.: Comparing the use of background
  knowledge by inductive logic programming systems. In: International Workshop
  on Inductive Logic Programming. pp. 199--230 (1995)

\bibitem{tao2004svm}
Tao, Q., Scott, S.D., Vinodchandran, N.V., Osugi, T.T.: Svm-based generalized
  multiple-instance learning via approximate box counting. In: International
  Conference on Machine Learning. p. 101 (2004)

\bibitem{tax2011bag}
Tax, D.M.J., Loog, M., Duin, R.P.W., Cheplygina, V., Lee, W.J.: Bag
  dissimilarities for multiple instance learning. In: Similarity-Based Pattern
  Recognition. pp. 222--234. Springer (2011)

\bibitem{wang2000solving}
Wang, J.: Solving the multiple-instance problem: A lazy learning approach. In:
  International Conference on Machine Learning. pp. 1119--1125 (2000)

\bibitem{zhang2005multiple}
Zhang, C., Platt, J.C., Viola, P.A.: Multiple instance boosting for object
  detection. In: Advances in neural information processing systems. pp.
  1417--1424 (2005)

\bibitem{zhang2001dd}
Zhang, Q., Goldman, S.A., et~al.: {EM-DD:} an improved multiple-instance
  learning technique. In: Advances in Neural Information Processing Systems.
  pp. 1073--1080 (2001)

\bibitem{zhou2005multi}
Zhou, Z.H., Jiang, K., Li, M.: Multi-instance learning based web mining.
  Applied Intelligence  22(2),  135--147 (2005)

\bibitem{zhou2009multi}
Zhou, Z.H., Sun, Y.Y., Li, Y.F.: Multi-instance learning by treating instances
  as non-iid samples. In: International Conference on Machine Learning. pp.
  1249--1256 (2009)

\end{thebibliography}

\end{document}